# Solving Scene Understanding for Autonomous Navigation in Unstructured Environments

Naveen Mathews Renji, Kruthika K,Manasa Keshavamurthy, Pooja Kumari, S. Rajarajeswari
Department of Computer Science Engineering
M.S. Ramaiah Institute of Technology
Affiliated to Visvesvaraya Technological University
Bangalore, India.

*Abstract —  Autonomous vehicles are the next revolution in the automobile industry and they are expected to revolutionize the future of transportation. Understanding the scenario in which the autonomous vehicle will operate is critical for its competent functioning. Deep Learning has played a massive role in the progress that has been made till date. Semantic Segmentation, the process of annotating every pixel of an image with an object class, is one crucial part of this scene comprehension using Deep Learning. It is especially useful in Autonomous Driving Research as it requires comprehension of drivable and non-drivable areas, roadside objects and the like. In this paper semantic segmentation has been performed on the Indian Driving Dataset which has been recently compiled on the urban and rural roads of Bengaluru and Hyderabad. This dataset is more challenging compared to other datasets like Cityscapes, since it is based on unstructured driving environments. It has a four level hierarchy and in this paper segmentation has been performed on the first level. Five different models have been trained and their performance has been compared using the Mean Intersection over Union. These are UNET, UNET+RESNET50, DeepLabsV3, PSPNet and SegNet. The highest MIOU of 0.6496 has been achieved. The paper discusses the dataset, exploratory data analysis, preparation, implementation of the five models and studies the performance and compares the results achieved in the process.*

*Keywords — Image Processing, Semantic Segmentation, Machine Learning (ML), Convolutional Neural Networks (CNN) U-Net, PSPNET, DeepLabV3, Segnet, MIoU, Deep Learning (DL)*

## I THE INTRODUCTION

Almost every automobile company is investing time, money and human resources into research that aims to bring the world closer to autonomous driving. The state-of-the-art models that were used on the structured roads fail to perform well with the Indian roads. A lot of research is still in its premature phase as for this kind of research, large datasets are required and though there are many available for structured road environments that have well-defined traffic lanes, adherence to traffic rules and organised vehicular and pedestrian movement, there was no such dataset available for high variant background images of the Indian roads Indian roads until recently.

This project uses Mean Intersection over Union metric for measure of accuracy and efficiency of each model. For each class (except 26), the  true positives, false negatives and false positives, (TP, FN, FP respectively) are calculated and all the prediction maps and ground truths will be resized to 720p using nearest neighbour algorithm over the whole  of the test split of the dataset. The formula TP/(TP+FN+FP) is used to compute the IoU (Intersection Over Union) for each class and their mean value is considered as the MIoU score which is used as the metric for the segmentation challenge.

The state-of-the-art methods which have worked well on such structured environment datasets like Cityscapes do not tend to perform well on the recently compiled Indian Driving Dataset which is based on roads of Hyderabad and Bengaluru. In this project the objective was to try to improve semantic segmentation on this dataset in an attempt to solve the autonomous driving challenges on Indian roads.

The aim was to develop a semantic segmentation model that can precisely predict obstacles and objects on the India roadway dataset that may enable autonomous driving on Indian roads. Further research into what can help improve the accuracy of such models and a study towards how far mankind are from achieving autonomy over drivers from Indian roads and get to a stage where autopilot is a practical deliverability on Indian roads.

Previously available datasets focused on structured driving environments like well-defined traffic lanes, adherence to traffic rules and organised vehicular and pedestrian movement which is not the case in countries like India. This dataset contains 10,004 images that are annotated with 34 classes which were collected from exactly 182 drive sequences on Bangalore and Hyderabad roads. This dataset can be used to make autonomous navigation a possibility in difficult environments. There are multiple challenges due to which the state-of-the-art models that performed well on other structured roads cannot perform similarly on indian roads. These include large intra-class appearance variations like trucks, cars, two and three wheelers, old and new vehicles, presence of novel road objects, unmarked and damaged roadways, varied and harsh ambient conditions (rainy, sunny), high density and unpredictability of traffic. Compared to existing datasets for navigation, this dataset has more number of labels and traffic participants. The dataset was benchmarked using the DRN-D-38 and the ERFNet model.

## II    RELATED WORKS

A literature survey on semantic segmentation and weakly supervised object detection explored different approaches and algorithms, with each method trying to solve the problem of predicting the image labels in their own way. Many papers suggest semantic segmentation as it is a supervised learning model which can give more accuracy. Some recommend the weakly supervised model for its ability to handle images that are very variant from the training dataset images.

In [1], the authors have used multi-scale inference. To improve the results of semantic segmentation multiple image scales are passed through a network and then the results are combined with averaging or max pooling. The authors have used an attention-based approach to combine multi-scale predictions. Predictions at certain scales are better at resolving particular failure modes, and the network learns to favour those scales for such cases in order to generate better predictions. They have used a hierarchical attention mechanism for the same which reduces the memory usage by four times to train when compared to other recent approaches as it requires augmenting the training pipeline with only one extra scale in contrast to others. Through this the network learns to predict a relative weighting between adjacent scales. This also allows training with larger crop sizes which leads to greater model accuracy. In addition to this, at the inference time scales can be selected flexibly. They also used a hard threshold based auto-labelling strategy of coarse images to increase variance and thereby to improve generalization. They have worked on the Cityscapes and Mapillary Vistas datasets and achieved 85.1 and 61.1 IOU values respectively. They have the highest IOU in the Cityscapes dataset which is closest to the IDD dataset.

In [2], the authors solve the problem of context aggregation in semantic segmentation. Considering the label of a pixel as the category of the object that the pixel belongs to, the authors have used object-contextual representations, characterizing a pixel by exploiting the representation of the corresponding object class. First, the object regions were learnt under the supervision of the ground-truth segmentation the object region representation was computed by aggregating the representations of the pixels lying in the object region and lastly the relation between each pixel and each object region was computed, and the representation of each pixel was augmented with the object-contextual representation which is a weighted aggregation of all the object region representations. The proposed model performed well on various benchmarks like Cityscapes, ADE20K, LIP, PASCAL-Context and COCO-Stuff.

In [3], the authors have worked on panoptic segmentation which includes both semantic and instance segmentation. The Efficient Panoptic Segmentation (EfficientPS) architecture was introduced that incorporates a shared backbone with a new feature aligning semantic head, a new variant of Mask R-CNN as the instance head, and a novel adaptive panoptic fusion module. The new panoptic backbone consisted of an augmented EfficientNet architecture, and their proposed 2-way FPN that both encodes and aggregates semantically rich multiscale features in a bidirectional manner. In addition they incorporated a new semantic head that aggregates fine and contextual features coherently before fusion for better object boundary refinement and a new variant of Mask R-CNN as the instance head. They also propose a novel panoptic fusion module that dynamically adapts the fusion of logits from the semantic and instance heads based on their mask confidences and concurrently integrates instance specific 'stuff' classes with 'thing' classes to perform the panoptic prediction. They also contributed to the KITTI panoptic segmentation dataset that provides panoptic groundtruth annotations for images from the challenging KITTI benchmark dataset. Extensive evaluations on Cityscapes, KITTI, Mapillary Vistas and Indian Driving Dataset demonstrate that this proposed architecture has performed well on all these four benchmarks and is one of the most efficient and fast panoptic segmentation architectures to date.

In [4], The authors introduced Panoptic-DeepLab, a simple, strong, and fast system for panoptic segmentation that uses bottom-up method and achieves comparable performance of two-stage methods and thus yields fast inference speed. It adopts the dual-ASPP and dual-decoder structures specific to semantic and instance segmentation. The semantic segmentation follows typical design of any semantic segmentation model like DeepLab and the instance segmentation involves a simple instance center regression and is class agnostic. As a result, the single Panoptic-DeepLab simultaneously performs well at all three Cityscapes benchmarks. Additionally, since it is equipped with MobileNetV3, it runs nearly in real-time with a single

1025 X 2049 image (15.8 frames per second), and achieves competitive performance on Cityscapes (54.1 PQ% on test set). Its performance is on par with several other top-down approaches on one of the most challenging COCO dataset. They strongly demonstrated that a bottom-up approach could also deliver state-of-the-art results on panoptic segmentation.

Pre-training is a dominant paradigm in computer vision. This can be validated from the fact that supervised ImageNet pre-training is commonly used in the initialization of the backbones of object detection and segmentation models. In [5], however, the authors demonstrate that ImageNet pre-training has limited impact on COCO object detection. Here, self-training is used as another method to utilize additional data on the same setup and contrast it against ImageNet pre-training which is further investigated. The study not only reveals the generality and flexibility of self-training but also these additional insights: 1) stronger data augmentation and more labeled data further diminish the value of pre-training, 2) unlike pre-training, self-training proves extremely helpful when stronger data augmentation is used, in both high-data and low-data regimes, and 3) in case where pre-training is helpful, self-training further improves upon pre-training. For example, on the COCO object detection dataset, pre-training improves accuracy when one fifth of the labeled data is used, and deteriorates accuracy when all of the labeled data is used. Self-training, on the other hand, shows positive improvements from +1.3 to +3.4AP across all dataset sizes. In other words, self-training works remarkably well on the same setup on which pre-training does not (using ImageNet to help COCO). On the PASCAL segmentation dataset, which is much smaller than COCO, pre-training does help significantly, while self-training improves upon the pre-trained model. On COCO object detection, 54.3AP was achieved which is an improvement of +1.5AP over the strongest SpineNet model. On PASCAL segmentation, 90.5 MIOU was achieved which is an improvement of +1.5% MIOU over the previous state-of-the-art result achieved by DeepLabv3+.

The literature survey revealed that for every purpose a different model is required when it comes to image analysis for object detection. In the cases that require prediction based on similar objects and driving environment, a strictly supervised learning is useful. But for the problem addressed in this paper, a model that can make precise and accurate predictions given unclear and noisy images that may vary a lot from the training data, is required. Hierarchical attention-based approaches to combine multi-scale predictions, dual-ASPP and dual-decoder structures, constrained Lagrangian dual optimization and CNN segmentation have shown much promise to the resolution of our problem.

## III    METHODOLOGY

We start by defining the problem statement for our project followed by gathering datasets, data analysis, preprocessing, preparation, training and evaluation and finally prediction is done at pixel level and performance is analysed. Figure 1 depicts the flow of data and operations followed in our project.

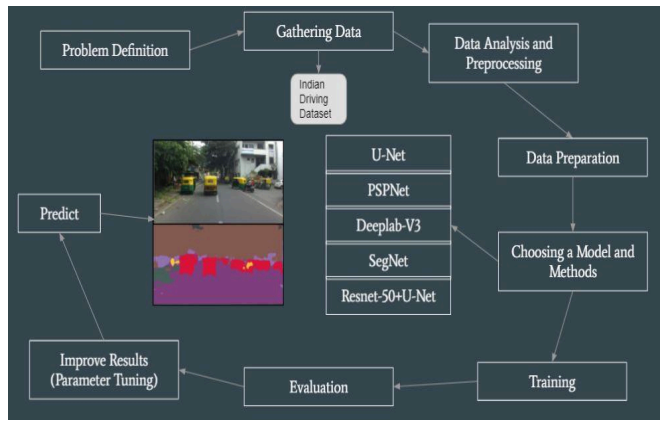

Figure 1: Flow Diagram

### 3.1 DATASET GATHERING

Many Datasets have become available for research and study on Autonomous navigation, but they have all been focused upon structured roadways only. The Dataset India Driving Dataset is collected here http://idd.insaan.iiit.ac.in from AutoNUE Challenge 2019 on Image segmentation. The front-facing camera attached to the vehicle that drove around both urban and rural areas of Hyderabad and Bangalore is responsible for clicking all the images in the dataset. Though most of the images are of Full HD (1080p), there are a few of 720p and other resolutions.

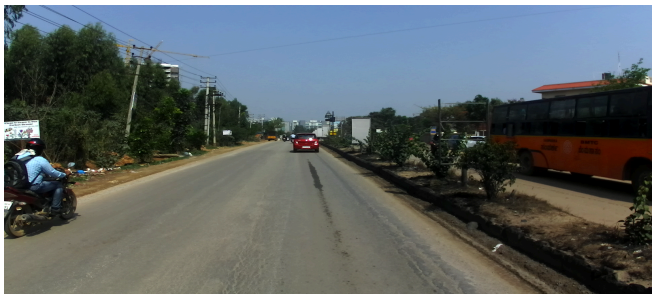

Figure 2: Image Depicting the IDD.

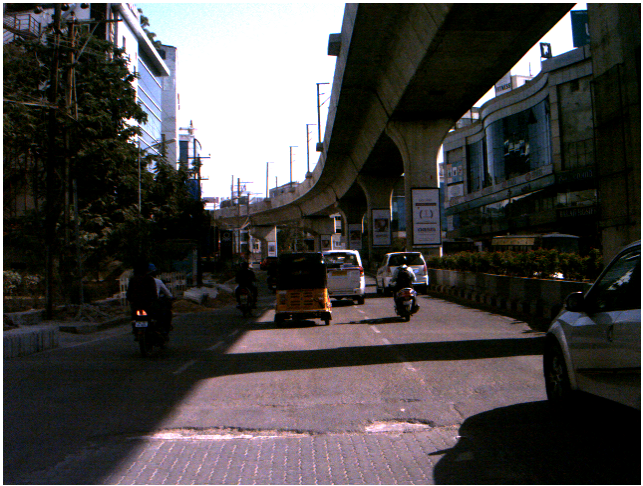

Figure 3: Image Depicting the IDD.

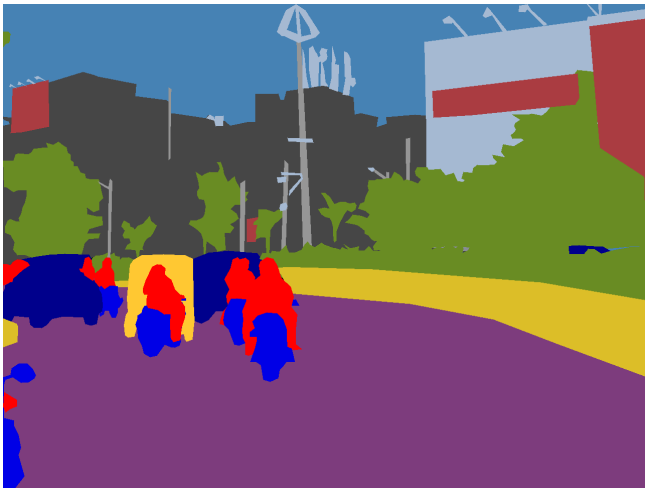

Figure 4: Labelled Image Depicting the IDD images.

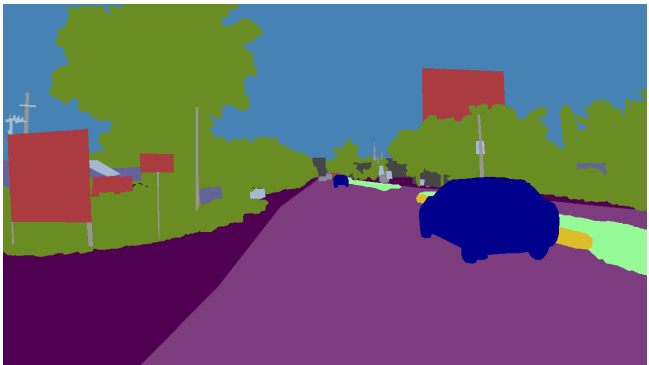

Figure 5: Labelled Image Depicting the IDD images.

### 3.2. DATA ANALYSIS AND PREPROCESSING

20,000 images that make up a total of 30 GB of memory collected from 182 rounds of driving and annotated finely with 34 classes from Indian roads make up the whole Indian Driving Dataset. Images of roads from the cities and outskirts of Hyderabad and Bangalore India have been taken to create this dataset. Various scenarios have been mixed up in the dataset. Images containing single lanes and double lanes, highways and roads with and without traffic of varying levels in both urban and rural areas of Bangalore and Hyderabad.. Data consists of a four-level label hierarchy where the highest hierarchy of labels consists of 7 class labels as Drivable, NonDrivable, Living Things, Vehicles, Roadside Objects, Far Objects, Sky.

The Image and Label mask both have a resolution of 1920 x 1080 (width x height) and the Total number of data samples are considered to be distributed into Train, Validation, Test as 14027, 2036, 4038 respectively. This step performs Image Data Analysis, Label Data Analysis, Comparison with Cityscapes Dataset and Analysing Data Statistics. Preprocessing also includes simplifying the file structure of the Dataset and making it simple for I/O operations.

### 3.3 RISK IDENTIFICATION

Risks can be classified as:
- Wrong recognition of obstacles
- Improper detection of pedestrians
- Missing frames due to improper data transfer
- Insufficient Training
- Incorrect training data
- Incorrect lighting or angle issue
- Unseen obstacles and vehicular movement

### 3.4 FUNCTIONAL REQUIREMENTS

Model should be :
- Invariant to noise in the background.
- Insensitive to colors and tones.
- Insensitive to lighting conditions.
- Works for a reasonable navigation range of 40 metres.
- Precisely identify pedestrians and animals
- Accurately detect vehicles and direction of movement
- detect road navigation obstacles

### 3.5 DATA PREPARATION

Image and Label Data Preparation Includes:
- Reading all images and Labels one after another from Directory.
- Resizing Images and Labels to same height and width
- Normalizing the pixel values in an image by dividing by 255
- Performing one hot encoding on mask resulting in a 3D matrix
- Saving Prepared Data of Image and Label in sparse representation for future Usage

- Shuffling prepared Data Samples and performing Train Test split on Data
- Implementing a Data generator to generate prepared Data Samples during Training.

### 3.6. CHOOSING A MODEL.

There are many Deep learning Architecture that researchers have created over the years but this case study includes the implementation of only a bunch of algorithms. The Deep Learning model architectures used are below :
- U-Net
- PSPNET
- DeepLabV3
- Segnet

### 3.7. TRAINING

This is considered the bulk of machine learning which refers to building a learning model. The train-test split consists of 60-15-25 for training, validation, and testing respectively. This is used for training the deep learning model. Training can be done in two ways either by generating samples by reading and preprocessing data for each batch or by just starting training with already saved preprocessed data of Dataset. The Image resolution is reduced while training from 1920 x 1080 to 480 x 240 (width x height) because of the limited availability of hardware resources.

### 3.8. EVALUATION

The model is evaluated and performance is measured. The goal of the Deep Learning model is to reduce Multi-class log loss. Used Mean Intersection over union as main metric to measure the performance of Deep Learning model used Accuracy and Confusion Matrix as metric of Reference

### 3.9. PARAMETER TUNING

Hyperparameter tuning is done to gain the best model possible. A bunch of Hyper Parameter values are tied and erred to increase MIOU as much as possible. Learning_rate_reduction and ModelCheckpoint methods are used to save the best model during training. Early stopping method is used to avoid overfitting and enhance results. The results of the model performance for this are Represented in the Box-strip plot and pretty table.

## IV DEEP LEARNING MODELS ANALYSIS

The project is separated into various modules, making it easier to work with and understand. We train U-Net, PSPNet, DeeplabV3, SegNet, Resnet-50 + U-Net models for semantic segmentation on IDD and improve accuracy of the models by tuning hyperparameters.

### 4.1 MODELS OBSERVATION

U-Net is an end-to-end fully convolutional network that performs a combination operation of information from both the downsampling and the upampling path thereby enabling a precise localization with the utilization of transposed convolutions and finally obtaining general information. It consists of only convolutional layers with no dense layers, because of which it can accept images of any size. The Deep learning model has misclassified some of the labels between Diving, Non-Driving and Roadside Object, Far Object. Transfer Learning with some variation on U-Net achieves good performance on Image segmentation when compared to Basic U-net.

DeeplabV3 chooses to preserve long-range context information and extract dense features in the network for better results. It solves the problem involving signal decimation and also learns multi-scale contextual features in the neural network. DeeplabV3 architecture achieves relatively good performance on Image segmentation when compared to other segmentation models.

PspNet is an effective network for complex scene understanding with a global pyramid pooling that uses contextual information from the image for semantic segmentation of the image. It has misclassified many of the labels due to fewer filters used while training because of limited availability of hardware. PspNet architecture achieves relatively lower performance on image segmentation when compared to other segmentation models.

SegNet is an Encoder-Decoder Network which is efficient with regards to space and time complexity. It focuses on the need to map low-resolution features to input resolution for Image Segmentation. The decoder network uses the max-pooling indices of feature maps to achieve a satisfied performance. The SegNet architecture achieves relatively good performance on Image segmentation when compared to other segmentation models.

Performance can be improved by training models with data in high resolution with more powerful hardware resources.

### 4.2 DETECTION OF LABELS

Red - Automobiles, Purple - Driveable area, Grey - Sky, Green - Non-Drivable, Yellow - Living things, Light purple - Roadside objects, Brown - Distant objects

- Chaotic Images - Ex. Images with garbage (Figure 6 (a) and (b)).

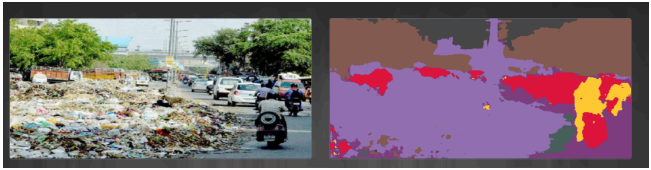

Figure 6 (a): Chaotic Image 1

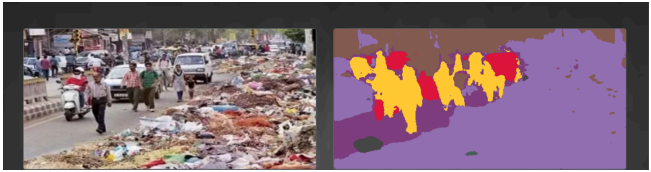

Figure 6 (b): Chaotic Image 2

- High Contrast Images - Ex. Images with shadows (Figure 7)

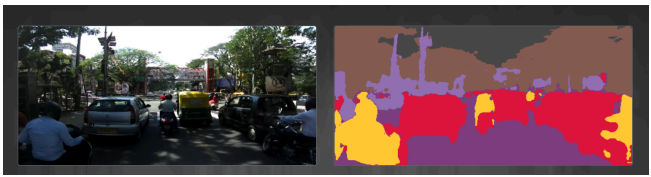

Figure 7: High Contrast Image

- Low Contrast Images - Ex. Images taken in rain (Figure 8)

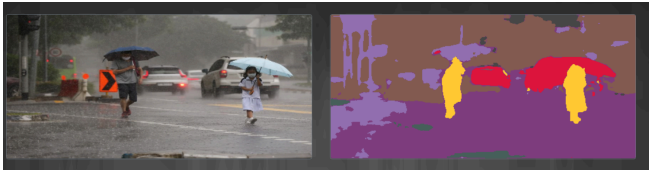

Figure 8: Low Contrast Image

- Balanced Scene - Traffic, People, garbage and Roadside objects (Figure 9 (a) and (b))

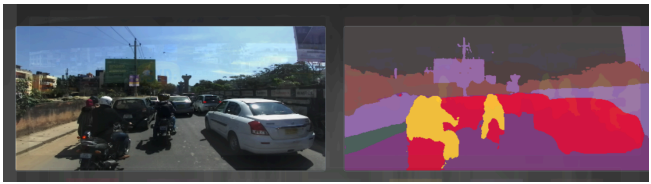

Figure 9 (a): Balanced Scene 1

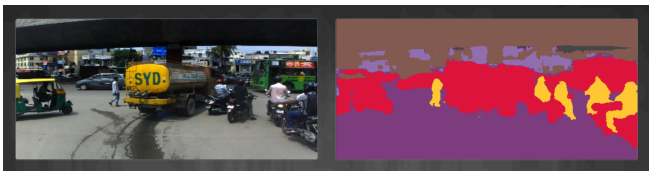

Figure 9 (b): Balanced Scene 2

## V  RESULTS

The following figures from 10 to 14 are of 3 images each. Image one represents the image captured from the camera on the vehicle. Image 2 represents the correct labels of the image pixels and Image 3 depicts the prediction of labels by each model.

1. UNET+RESNET50

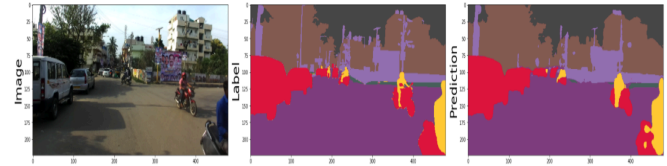

Figure 10: Results for UNET+RESNET50

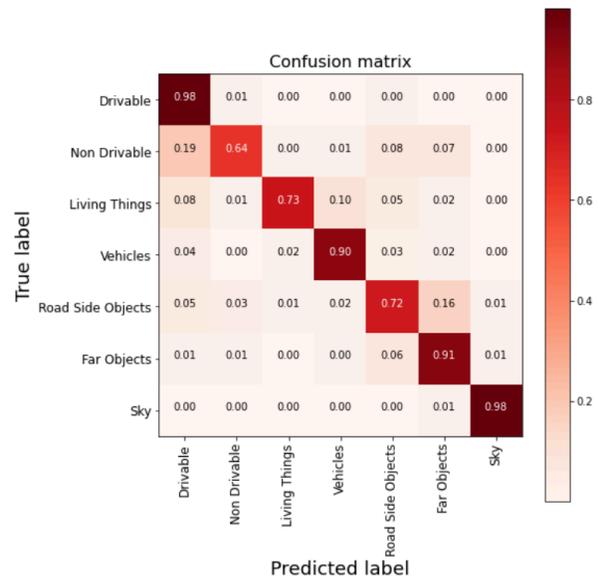

Figure 11: Confusion Matrix for UNET+RESNET50

It can be observed from the confusion matrix that the labels have been predicted with a fairly good level of precision by the model.

The two models, Unet and ResNet50+Unet were compared using their results and it was found that U-Net+ResNet50 performed better in comparison with U-net as shown in Tables 1 and 2.

| U-Net | MIOU | Accuracy |
|---|---|---|
| Train | 0.6642 | 0.9306 |

| Validation | 0.5843 | 0.8760 |
|---|---|---|
| Test | 0.5979 | 0.8888 |

Table 1: Results for UNET

| ResNet50+U-Net | MIOU | Accuracy |
|---|---|---|
| Train | 0.7487 | 0.9600 |
| Validation | 0.6389 | 0.9059 |
| Test | 0.6496 | 0.9160 |

Table 2 : Results for UNET+RESNET50

2. DEEPLABV3

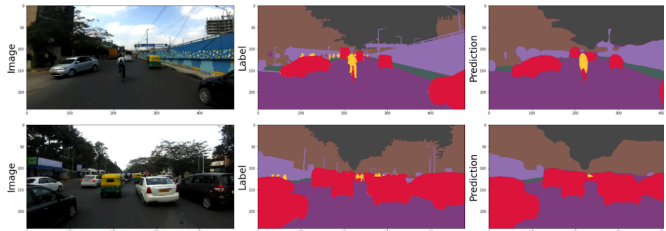

Figure 12: Results for DeepLabsV3

| MIOU Score | 0.6598 |
|---|---|
| Accuracy Score | 0.9279 |

Table 3: Results for DeepLabsV3

3. PSPNET

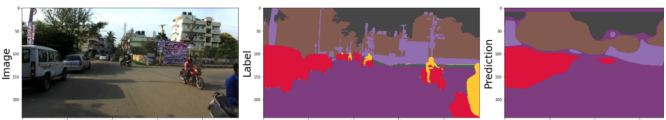

Figure 13: Results for PSPNet

| MIOU Score | 0.4284 |
|---|---|
| Accuracy Score | 0.7941 |

Table 4 : Results for PSPNet

4. SEGNET

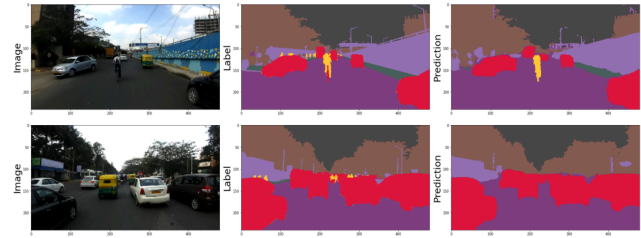

Figure 14: Results for SegNet

| MIOU Score | 0.6258 |
|---|---|
| Accuracy Score | 0.9154 |

Table 5 : Results for SegNet

6.1 COMPARISON OF RESULTS

The results of the model performance for this are Represented in the Box-strip plot.

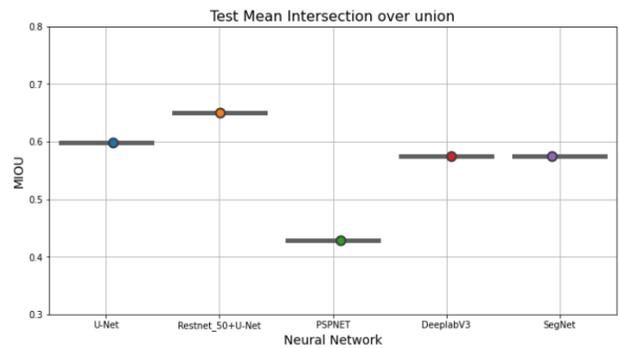

Figure 20: Box-Strip Plot for Comparison of Test Mean Intersection over Union of the Deep Learning Models

| Models | Train_MIOU | Train_Accuracy | Test_MIOU | Test_Accuracy |
|---|---|---|---|---|
| U-Net | 0.6642 | 0.9306 | 0.5979 | 0.8888 |
| Restnet50+U-Net | 0.7484 | 0.9600 | 0.6496 | 0.9160 |
| PSPNET | 0.4612 | 0.8225 | 0.4284 | 0.7941 |
| DeeplabV3 | 0.6598 | 0.9279 | 0.5752 | 0.8832 |
| SegNet | 0.6258 | 0.9154 | 0.5747 | 0.8814 |

Table 6 : Comparison Table for the Deep Learning Models

6.2 PERFORMANCE ANALYSIS

Semantic segmentation has many applications in image processing and computer vision and more accurate and faster methods have been emerging for image segmentation. All learning models above tend to have some confusion between Driving and Non-driving labels. The Altering of the Learning rate for some epochs improves MIOU and accuracy significantly. The Imbalanced Dataset can sometimes be a curse for performance improvement. Finally, the best performing model is Restnet_50+U-Net with highest MIOU value which performs better than other models.

## VI  CONCLUSION AND FUTURE SCOPE

Since the vitality of technological innovations in the field of algorithm efficacy and image processing will always remain atop, the possibilities are endless to the level of development possible in the image recognition and classification field. The fundamental aim of this project is to detect and label the various objects seen on Indian roads, making use of computational innovations in Machine Learning and Image Processing but our model of image processing can also be used for several other country's roadways. The future holds a lot of potential for the deployment of autonomous vehicles in the Indian roads as current technology is far from being able to navigate unmanned through the chaotic and haphazard unstructured road in the Indian subcontinent. Once a model that is powerful enough to classify and predict the movement of objects and roadways on the Indian territory, it will then become possible for a new age of autonomous travel over Indian roadways.